\documentclass[10pt,twocolumn,letterpaper]{article}

\usepackage[pagenumbers]{cvpr} 

\usepackage[dvipsnames]{xcolor}

\usepackage{graphics}

\definecolor{cvprblue}{rgb}{0.21,0.49,0.74}
\usepackage[pagebackref,breaklinks,colorlinks,citecolor=cvprblue]{hyperref}

\def\paperID{XXXX} 
\def\confName{CVPR}
\def\confYear{2024}

\title{Dubbing for Everyone:  Data-Efficient Visual Dubbing using Neural Rendering Priors}

\author{Jack Saunders\\
University of Bath \& Deepreel \\
{\tt\small https://jsaunders909.github.io/}
\and
Vinay P Namboordiri\\
University of Bath\\
{\tt\small https://vinaypn.github.io/}
}

\begin{document}

\twocolumn[{%
\renewcommand\twocolumn[1][]{#1}%
\maketitle
}]

\begin{abstract}

Visual dubbing is the process of generating lip motions of an actor in a video to synchronise with given audio. Recent advances have made progress towards this goal but have not been able to produce an approach suitable for mass adoption. Existing methods are split into either person-generic or person-specific models. Person-specific models produce results almost indistinguishable from reality but rely on long training times using large single-person datasets. Person-generic works have allowed for the visual dubbing of any video to any audio without further training, but these fail to capture the person-specific nuances and often suffer from visual artefacts. Our method, based on data-efficient neural rendering priors, overcomes the limitations of existing approaches. Our pipeline consists of learning a deferred neural rendering prior network and actor-specific adaptation using neural textures. This method allows for \textbf{high-quality visual dubbing with just a few seconds of data}, that enables video dubbing for any actor - from A-list celebrities to background actors. We show that we achieve state-of-the-art in terms of \textbf{visual quality} and \textbf{recognisability} both quantitatively, and qualitatively through two user studies. Our prior learning and adaptation method \textbf{generalises to limited data} better and is more\textbf{scalable} than existing person-specific models. Our experiments on real-world, limited data scenarios find that our model is preferred over all others. \footnote{Project page may be found at \url{https://dubbingforeveryone.github.io/}}

\end{abstract}    
\section{Introduction}
\label{sec:intro}

Dubbing is the process of translating video content from one language to another. Currently dubbing is performed on only the audio tracks, leaving the video unchanged. The result is a poor visual experience. More recently, visual dubbing has been proposed to make dubbed video content more appealing to audiences. Visual dubbing involves reconstructing the lip and mouth movements of the actor in a video to match new audio in a different language. When done correctly, Visual dubbing has the power to transform how global audiences will watch video content filmed in non-native languages, and allow content creators to reach more viewers across the world.

We argue that any successful video dubbing method must be \textcolor{blue}{\textbf{high-quality}}, \textcolor{ForestGreen}{\textbf{generalizable}}, \textcolor{orange}{\textbf{scalable}} and \textcolor{red}{\textbf{recognizable}}. It must be \textcolor{blue}{\textbf{high-quality}} so that consumers are not distracted by the synthesized lips, avoiding the 'uncanny valley' effect. This requires good video quality and good lip sync. It must be \textcolor{ForestGreen}{\textbf{generalizable}} in the sense that all actors, from A-list stars to background actors, should be dubbed effectively with as little as a few seconds of dialogue. It must be \textcolor{orange}{\textbf{scalable}}, a sufficiently large model may be high-quality and generalizable, but it will not be adopted if adding another actor to the model requires several days of training. Finally, an actor's style should be \textcolor{red}{\textbf{recognizable}} in the dubbed video. For instance, the actor's lips and teeth should look the same in the dubbed video as in the real one.      

\begin{figure}[t]
    \centering
    \includegraphics[width=\columnwidth]{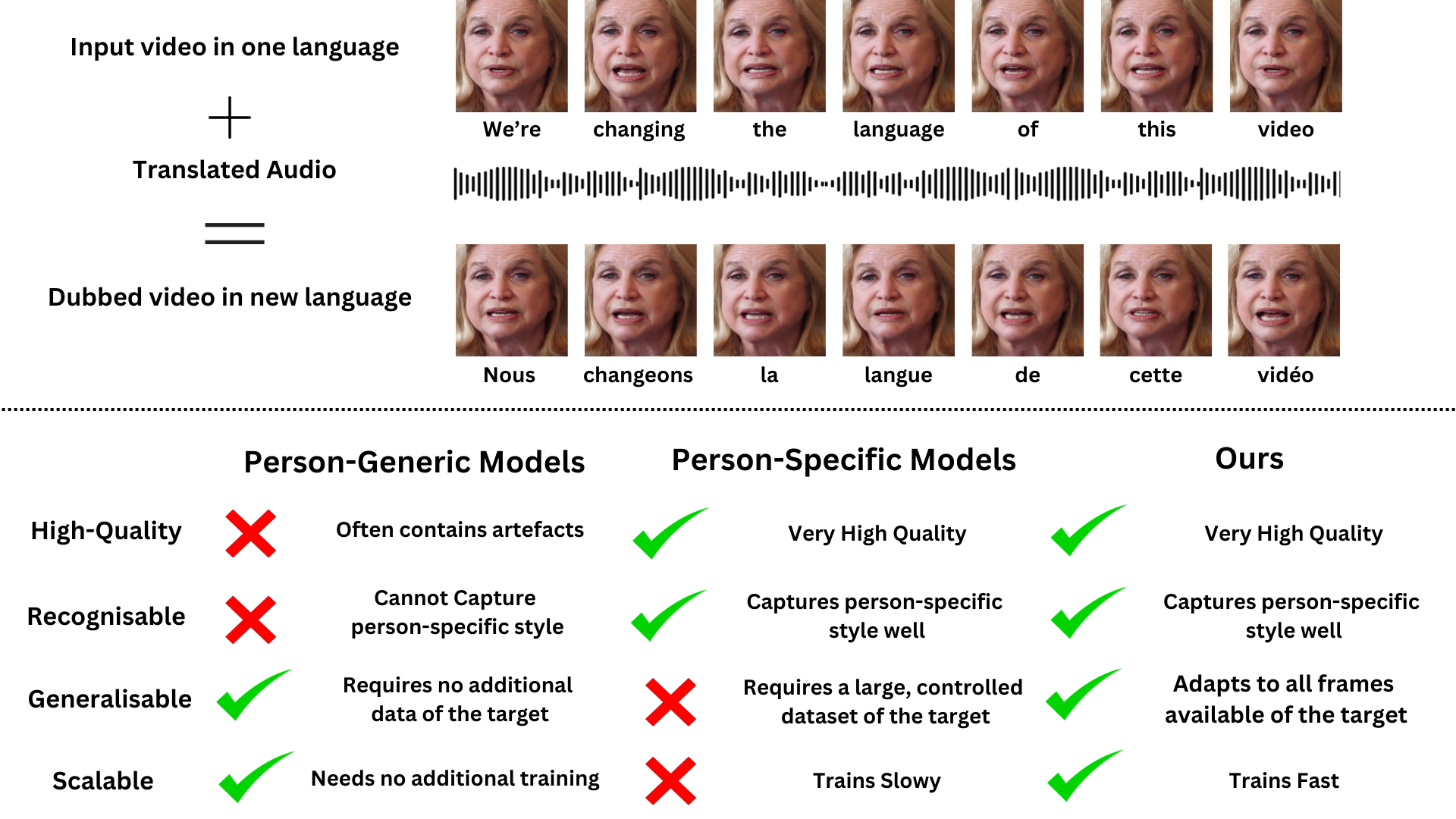}
    \caption{Our method, Dubbing for Everyone allows for the reconstruction of lip movements when dubbing video from one language to another. Our approach overcomes the limitations of person-generic and person-specific visual dubbing models, to produce a model that is of \textcolor{blue}{\textbf{high visual quality}}, \textcolor{red}{\textbf{recognisable}}, \textcolor{ForestGreen}{\textbf{generalisable}} and  \textcolor{orange}{\textbf{scalable}}}.
    \label{fig:title}
\end{figure}

Previous models are split between their focus on these criteria. We detail this in \cref{fig:title}. Some produce incredible quality video for a single actor under controlled conditions (e.g. \cite{thies2020nvp, Ye2023GeneFaceGA, tang2022radnerf}) such methods are high-quality and recognizable but will work only on the actor they are trained on. Others produce low-quality, but generalizable video (e.g. \cite{wang2023seeing, Wav2LipHQ, guan2023stylesync}. These methods can be applied to any audio and any video, but the outputs are rarely of good visual quality and they do not capture the style of the actors, instead providing generic lips that synchronize well with the audio.  

We propose Dubbing for Everyone. Taking the best of both approaches, we create a model that meets \textbf{\textit{all}} our criteria, allowing the high-quality dubbing of all actors, including those with short roles. We use a prior network that is trained across multiple actors and therefore can generalize, and also introduce actor-specific components that allow for our model to adapt to individuals and produce high-quality and recognizable dubbing. In particular, we adopt a multi-stage approach based on neural textures. We separate the task of visual dubbing into audio-to-lip-motion and video generation stages. This allows us to capture idiosyncratic motion and appearance on a per-subject basis explicitly. 

Our model is \textcolor{ForestGreen}{\textbf{generalisable}} due to the person-generic prior network training on a large dataset, we demonstrate this in \cref{tab:main_results} by using just 4 seconds of actor-specific data and further discuss this quality in \Cref{sec:dataset_size}. It is \textcolor{blue}{\textbf{high-quality}} due to the person-specific adaptation making effective use of all data, we achieve state-of-the-art in this respect as is shown in \Cref{tab:main_results}. Our method is \textcolor{orange}{\textbf{scalable}}, given a new identity, training is resumed from the prior network and is therefore much faster. We find an order of magnitude speedup compared to existing person-specific models (\Cref{sec:speed}). Person-specific appearance detail is captured in the neural textures, and idiosyncratic motion is captured due to the separation of the audio-to-motion component. This makes our method \textcolor{red}{\textbf{recognisable}} for any actor. We validate this through a user study in \Cref{tab:user_study}. The novel contributions in this paper may be summarised as:

\begin{itemize}
    \item{We present \textbf{Dubbing for Everyone}. A hybrid visual dubbing model using person-generic and person-specific components and capable of producing \textcolor{blue}{\textbf{high-quality}} and \textcolor{red}{\textbf{idiosyncratic}} results from just a few seconds of data.}
    \item{We train a prior deferred neural rendering network across many identities and learn actor-specific neural textures allowing us to adapt our model to new identities. The training of the prior network allows for \textcolor{ForestGreen}{\textbf{data-efficent dubbing}} resulting in a significant reduction in data requirements compared to existing person-specific models.}
    \item{We propose a novel post-processing algorithm that removes artefacts in the border around the generated video. This improves perceived quality (\Cref{tab:ablation}).}
    \item{We perform an extensive evaluation to show that our method achieves state-of-the-art for \textcolor{blue}{\textbf{quality}} (\Cref{tab:main_results} \& \Cref{tab:user_study}) and \textcolor{red}{\textbf{recognisability}} (\Cref{tab:main_results}), as well as being \textcolor{orange}{\textbf{an order of magnitude fast to train}} (\Cref{sec:speed} \& \Cref{tab:speed}) and \textcolor{ForestGreen}{\textbf{robust in few-shot scenarios.}} (\Cref{sec:dataset_size} \& \Cref{tab:main_results})}
\end{itemize}

\section{Related Work}
\label{sec:related}

Early works in visual dubbing \cite{VideoRewrite, Ezzat02} use a variety of methods. However, for the purposes of this work, we consider post-deep-learning models. Two separate classes of visual dubbing exist in this context: person-generic and person-specific models.

\subsection{Person Generic Models}

Person-generic models differ from person-specific models in that they can work zero-shot for any new person and for any audio. These methods typically are 2D-based.

One class of these models uses some form of expert discriminator to achieve lip sync. Early methods \cite{Wav2Lip, LipGAN} use an encoder-decoder model to predict frames from audio and use random reference frames of the same person. These methods use adversarial training combined with an expert Syncnet \cite{Chung16a} which predicts if video and audio are in or out of sync. A majority of person-generic models build upon this framework, but replace some of these components. Some seek to replace the encoder-decoder model with transformers \cite{Wang_2023_CVPR} or vector-quanitised models \cite{Wav2LipHQ}. Others change the type of expert discriminator \cite{wang2023seeing, Sun2022} or else replace the adversarial loss with a diffusion process \cite{shen2023difftalk, stypulkowski2022diffused}.

Another class of person-generic model works on the idea of flow. These models will estimate motion using either landmarks \cite{Yang:2020:MakeItTalk, ji2021audio-driven, kaisiyuan2020mead, Gururani_2023_ICCV} or pixel-based flow \cite{Siarohin_2019_NeurIPS}. This motion is then used to warp a reference image.

In either case, these models use only a single reference frame to encode the identity. This is a significant issue as a single image cannot contain enough information about either the appearance or talking style. For example, if the mouth is closed in the reference frame, it is not possible to predict what the mouth interior should look like. Our work, in contrast, is able to use all available frames of the target person for fine-tuning, enabling us to capture idiosyncratic qualities.

\begin{figure*}
    \centering
    \includegraphics[width=\textwidth]{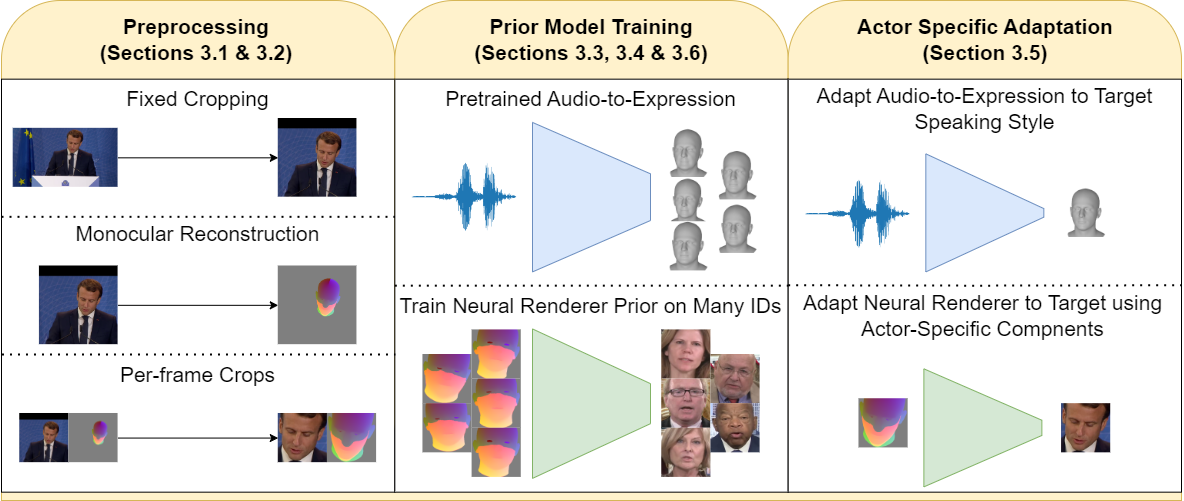}
    \caption{The pipeline of our method. We first apply preprocessing to our dataset (\Cref{sec:preprocess}) to obtain 3D reconstructions, tightly and stably cropped to the face. We next obtain person-generic audio-to-expression and neural rendering models using multiple subjects (\Cref{sec:generic}). Given a new subject, we then finetune both models for the given subject (\Cref{sec:finetune}).}
    \label{fig:method}
\end{figure*}

\subsection{Person Specific Models}

Person-specific models are trained per-person, usually under controlled conditions. As a result of this, they are typically much higher quality than person-generic models, but cannot produce results for anyone other than the person they were trained on. It is very common for person-specific models to use some form of 3D supervision in order to improve the results. By introducing 3D priors, certain characteristics such as the face shape or pose can be controlled for. 

One line of work builds upon the 3D Morphable Model \cite{3DMM2020, 3DMMOG}. The 3DMM allows for pose, lighting, shape and texture to remain constant, only changing the expression of the face. Some works achieve dubbing by having one actor provide the lip motions for another \cite{kim2018deep, NSPVD}, while others generate the lip motions from audio \cite{thies2020nvp, saunders2023read, AudioDVP, loy2020everybody, Ma_2023_AAAI}.

Another, more recent, line of research builds upon implicit 3D geometry. In particular, Neural Radiance Fields \cite{mildenhall2020nerf} (NeRFs) have been used to good effect for talking head generation. Audio-driven models \cite{guo2021adnerf, shen2022dfrf, tang2022radnerf, Ye2023GeneFaceGA} condition NeRFs on audio to produce free-viewpoint renderings.

Several other works have also addressed the problem from a person-specific viewpoint and are less easily grouped. Models range from using a diffusion auto-encoder \cite{du2023daetalker} to person-specific landmark-based models \cite{lu2021live, SynthObama}. 

While ranging in the methodology, all person-specific models share some important qualities. They all produce very high-quality output but come with significant data requirements ranging from about 15 seconds \cite{shen2022dfrf} to upwards of 10 minutes \cite{du2023daetalker}. In contrast, our method is able to achieve similar quality using as little as 4 seconds of training data, thanks to our person-generic prior network training and person-specific adaptation.
\section{Method}

Our method builds upon deferred neural rendering \cite{thies2019deferred}. The key to our method is the training of a prior deferred neural rendering network (\Cref{sec:generic}) which is person-generic, and the adaptation to new actors using neural textures (\Cref{sec:finetune}). This method (\Cref{fig:method}) requires an order of magnitude less data than previous neural textures approaches. Before training the prior network, we run a  preprocessing stage, which involves cropping the video frames to the face region and performing monocular reconstruction (\Cref{sec:monocular}) to get a parameterized 3D representation of the face.

Using an existing speech-driven animation model \cite{Thambiraja_2023_ICCV}, we are able to control the 3D model and in turn alter the lip motions of a given video (\Cref{sec:a2e}). We find that some artefacts are left during the video generation process, so we propose a postprocessing step that is able to remove these (\Cref{sec:post}).

\begin{figure*}
    \centering
    \includegraphics[width=\textwidth]{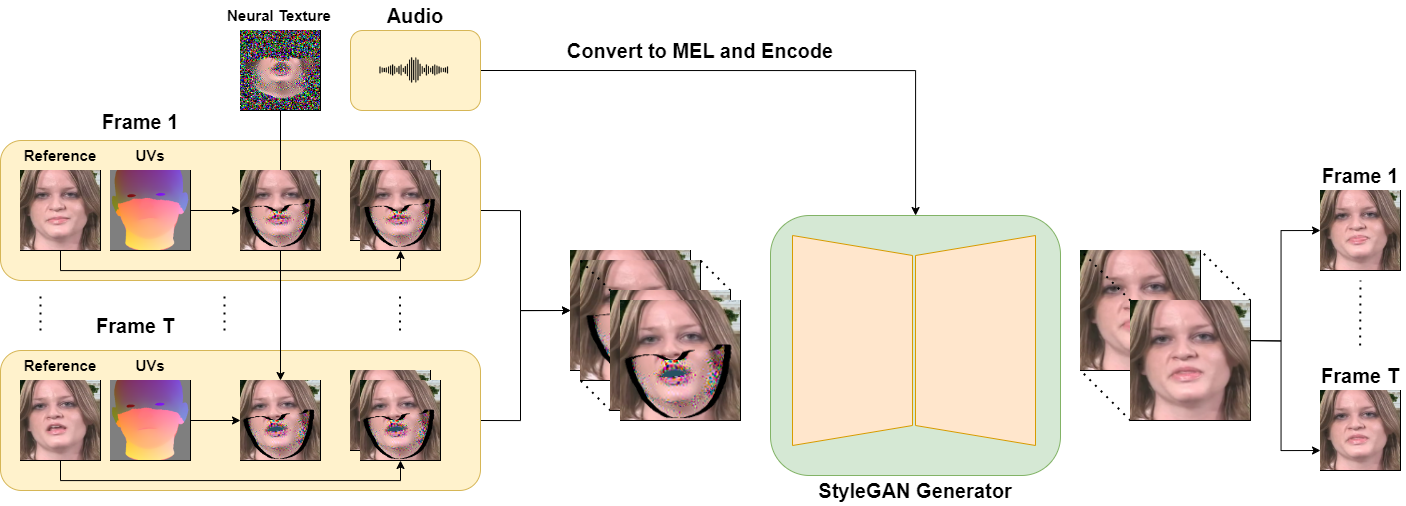}
    \caption{The architecture of our model. We take the UV rasterisations of each frame in a window of length T and sample a neural texture. Using a mask, we combine this with the real frame (\Cref{fig:NetworkInput}). These are concatenated with random reference frames of the same person and stacked across the channel dimension. A StyleGAN2-based \cite{guan2023stylesync, Karras2019stylegan2} generator is then used to convert these into T photorealistic frames.}
    \label{fig:arch}
\end{figure*}

\begin{figure}[!h]
    \centering
    \includegraphics[width=\columnwidth]{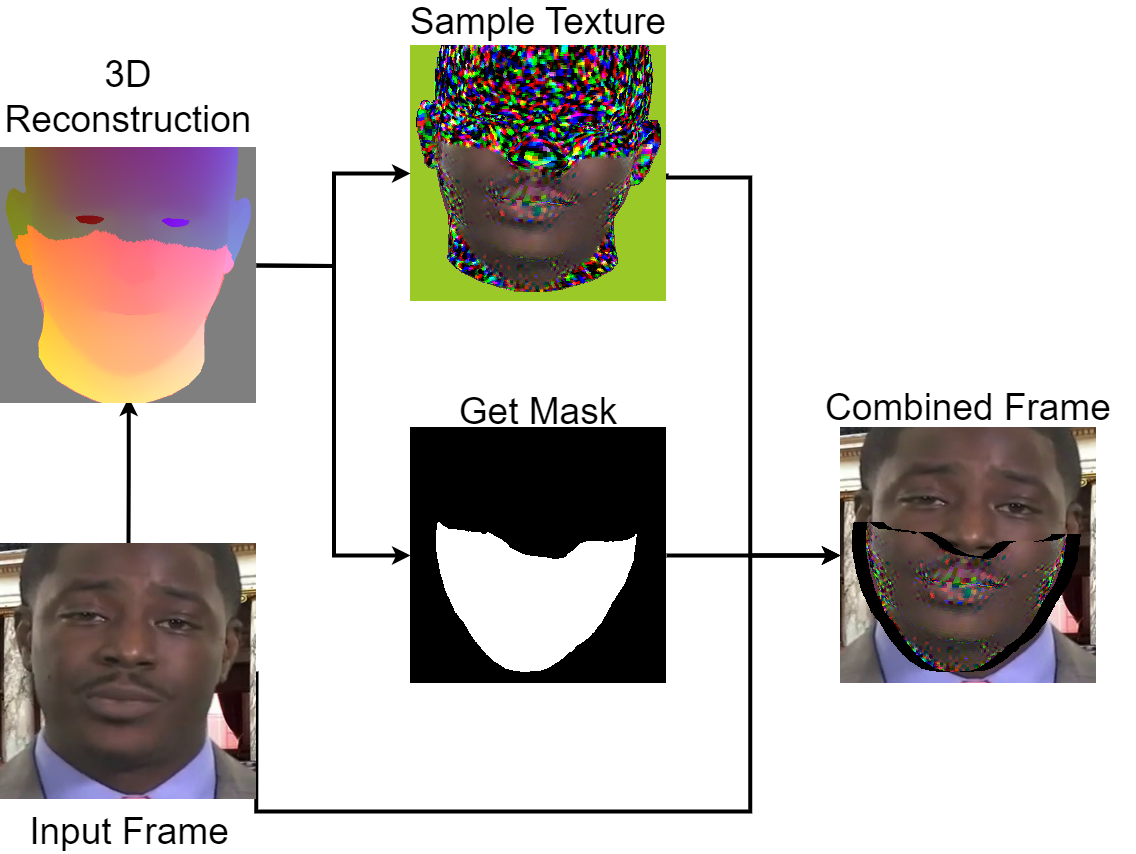}
    \caption{The input to the image-to-image network. We sample the neural texture onto the rasterised mesh and use the mesh to estimate a mask. The input is computed using this mask, the target frame, and the rasterised texture.}
    \label{fig:NetworkInput}
\end{figure}

\subsection{Monocular Reconstruction}
\label{sec:monocular}

Following similar 3DMM-based neural texture approaches \cite{thies2020nvp, thies2019deferred}, we first fit a 3DMM to each frame of the ground truth videos using differentiable rendering. For the 3DMM, we use FLAME \cite{FLAME:SiggraphAsia2017}. FLAME uses a combination of linear blendshapes and blend skinning to control a full-face rig with 5023 vertices. We use a three-stage process for this fitting. We cover more detail in the supplementary material.

\noindent \textbf{Stage 1:} First we estimate the shape parameters of the FLAME model using MICA \cite{MICA:ECCV2022}. MICA predicts the shape parameters from a single frame and is shown to be very accurate. We then fix the shape parameters.

\noindent \textbf{Stage 2:} Now, with the shape fixed, we optimise other non-varying parameters by jointly optimising a regularised photometric loss function over several frames at the same time. We then fix the texture and camera parameters, in addition to the shape.

\noindent \textbf{Stage 3:} Finally we optimise the variable parameters for each frame. These are expression, pose and lighting. We initialise parameters for frame $t$ using the parameters for frame $t-1$.

\subsection{Preprocessing}
\label{sec:preprocess}

We next crop the face to $256 \times 256$ pixels. We find that pretrained face detectors (e.g. \cite{MediaPipe, Deng2020CVPR}) suffer from two issues. The first is jitter between frames and the second is the bounding box varies based on the jaw pose. To overcome both these issues we use our tracking data to generate bounding boxes. We render the vertices of the meshes to obtain in \Cref{sec:monocular} with the jaw in several positions and then find the bounding box that contains all of these positions. We then add a small margin to get the final box. More details can be found in the supplementary material.

We also use the tracked data to generate masks. This means that we only need to predict the necessary pixels as opposed to existing methods which predict the entire lower half of the face. We rasterise a predefined mouth texture mask which may be seen in \Cref{fig:NetworkInput}.

\subsection{Architecture}
\label{sec:arch}

In this section, we describe the architecture of our model. Inspired by previous work \cite{thies2019deferred}, we adopt a deferred neural rendering approach using neural textures. The model contains two components: learnable neural textures which are similar to standard RGB, uv-based, texture images with high-dimensional feature vectors; and a deferred neural renderer, an image-to-image network that takes these rasterised neural features and converts them into realistic video (\Cref{fig:arch}). 

Existing models require several minutes of training data. The primary novelty of our work is in adapting this method to work few-shot, using only a small dataset of a given actor. A key insight to obtaining this is to note that much of the person-specific information can be stored in the neural textures, allowing the image-to-image network to be generalised across multiple subjects. In order to help the image-to-image network generalise, we use a reference frame as is done in person-generic works \cite{Wav2Lip, Wav2LipHQ, guan2023stylesync, shen2023difftalk, stypulkowski2022diffused}.

In order to improve the quality of the generations, we replace the UNET used in previous works with a modified version of the StyleGAN2 \cite{Karras2019stylegan2} architecture used in StyleSync \cite{guan2023stylesync}. Instead of providing masked target frames as is done in StyleSync, we mask out the target frame using the rasterization of a predefined mask on the 3DMM (\Cref{sec:preprocess}), and fill in the masked regions with the rasterized neural texels. This can be seen in \Cref{fig:NetworkInput}.

Using the generator architecture from StyleSync also allows us to condition the video generation on audio. To improve temporal stability, we provide the generator with access to a window of frames surrounding the target, and predict the same window of final video. The architecture is best understood by viewing \Cref{fig:arch} and referring to the StyleSync paper \cite{guan2023stylesync}.

\subsection{Training the Prior Model}
\label{sec:generic}

To train the prior deferred neural rendering network model, we use multiple identities. The network weights are shared for all identities, but we have a different randomly initialised neural texture for each. We jointly optimise the network and textures to minimise the following loss:

\begin{equation}
    \mathcal{L} = \lambda_1 \mathcal{L}_1 + \lambda_{\text{VGG}} \mathcal{L}_{\text{VGG}} + \lambda_{\text{reg}} \mathcal{L}_{\text{reg}} + \lambda_{\text{adv}} \mathcal{L}_{\text{adv}}
\end{equation}

$\mathcal{L}_{1}$ is a simple $\ell_1$ loss computed between the generated window of frames and the ground truth. In order to encourage the network to produce better results in the lower face and mouth region, we compute masks for these areas and weigh them higher. Specifically, we weight the mouth region at 10 times the background and upper face, and the lower face region at 8 times the background.

$\mathcal{L}_{\text{VGG}}$ is a VGG-based \cite{johnson2016perceptual} style loss. This is computed using a pre-trained VGG network and is calculated for each frame in the window individually, taking the mean. This is a perceptual loss and is known to improve image quality.

$\mathcal{L}_{\text{adv}}$ is an adversarial loss. We jointly train the model with a discriminator and use an LSGAN \cite{mao2017least} formulation for the adversarial loss. The discriminator is identical to the one used in StyleSync \cite{guan2023stylesync} but it is shown all frames in the window to encourage temporal consistency as demonstrated in previous works \cite{Wav2Lip}.

Finally, $\mathcal{L}_{\text{reg}}$ is a regularisation loss for the neural textures, following previous work \cite{thies2019deferred}. It is computed as the $\ell_1$ distance between the first three channels of the rasterised texture and the target frame. This encourages the first three channels of the texture to mimic a standard, diffuse, RGB texture.

\subsection{Adapting to New Identity}
\label{sec:finetune}

Given a new actor, we adapt our model to them. To do this, we first initialise a new random neural texture and use the deferred neural rendering prior network from \cref{sec:generic}. We train the texture from scratch but use the prior network as initialisation for the deferred neural renderer. To help the model maintain its generalisation, we also include data from other identities from the training set of the generic model. Specifically, we include person-generic training data in the person-specific dataset at a ratio of 1:1. We refer to this as a \textbf{mixed training strategy}. The mixed training strategy allows the deferred neural rendering network to continue learning from a wide data distribution, including, for example, poses that may not be in the actor-specific dataset.

\subsection{Audio-to-Expression Model}
\label{sec:a2e}

Given our neural rendering model, we are able to convert rasterizations of the 3DMM to realistic video. In order to change the lip motions of the video, we need to alter the model's expression parameters. To do this, we make use of state-of-the-art speech-driven animation models. In particular, we use Imitator \cite{Thambiraja_2023_ICCV}. Imitator is a transformer-based model that allows for speaking style adaptation. We use the pre-trained Imitator model. In order to adapt to the speaking style of the new individual, we add a final layer to the network which applies a linear transformation independently for each expression and jaw pose parameter. We then train only this layer for each individual.

\begin{table*}
  \centering
  \resizebox{\textwidth}{!}{%
  \begin{tabular}{c|cccccc|cccccc}

      & \multicolumn{6}{c|}{\textcolor{ForestGreen}{\textbf{100 Frames of Training Data}}} & \multicolumn{6}{c}{1000 Frames of Training Data} \\
     Method & \textcolor{blue}{\textit{PSNR}} $\uparrow$ & \textcolor{blue}{\textit{SSIM}} $\uparrow$ & \textcolor{blue}{\textit{FID}} $\downarrow$ & \textcolor{blue}{\textbf{Qual}} $\uparrow$ &  \textcolor{blue}{\textbf{Lip}} & $\uparrow$ \textcolor{red}{\textbf{ID}} $\uparrow$  & \textcolor{blue}{\textit{PSNR}} $\uparrow$ & \textcolor{blue}{\textit{SSIM}} $\uparrow$ & \textcolor{blue}{\textit{FID}} $\downarrow$ & \textcolor{blue}{\textbf{Qual}} $\uparrow$  & \textcolor{blue}{\textbf{Lip}} $\uparrow$ & \textcolor{red}{\textbf{ID}} $\uparrow$ \\

     \midrule

    Wav2LipHQ [WACV23] & 27.70 & 0.895 & 6.78 & 3.53 & 3.67 & 3.10 & 27.70 & 0.895 & 6.78 & 3.53 & 3.67 & 3.10 \\
    StyleSync [CVPR23] & \textbf{29.26} & 0.899 & 7.07 & \textbf{3.77} & 3.50 & 3.20 & 29.26 & 0.899 & 7.07 & 3.77 & 3.50 & 3.20 \\
    TalkLip   [CVPR23] & 28.34 & 0.887 & 9.98 & 2.47 & 3.59 & 3.03 & 28.34 & 0.887 & 9.98 & 2.47 & 3.59 & 3.03 \\
    \midrule
    RAD-NeRF [Preprint]& 22.63 & 0.732 & 30.63 & 2.17 & 1.67 & 2.75 & 26.23 & 0.835 & 22.94 & 2.33 & 1.67 & 3 \\
    GeneFace [ICLR23]& 21.87 & 0.720 & 38.50 & 2.00 & 1.92 & 2.77 & 24.43 & 0.797 & 26.38 & 2.67 & 2.75 & 3.08 \\
    \bottomrule
    Ours Baseline & 27.92 & 0.888 & 10.52 & 2.90 & 3.33 & 3.13 & 29.00 & 0.899 & 8.61 & 3.06 & 3.23 & 3.10 \\
    Ours Texture Only & 28.30 & 0.891 & 10.60 & 2.80 & 3.13 & 2.97 & 29.00 & 0.900 & 8.47 & 3.30 & 3.60 & 3.57 \\
    Ours Full & 29.10 & \textbf{0.899} & \textbf{5.76} & 3.57 & \textbf{3.77} & \textbf{3.83} & \textbf{29.30} & \textbf{0.904} & \textbf{5.44} & \textbf{3.90} & \textbf{3.80} & \textbf{4.03} \\
    \hline
     Real & 100.00 & 1.00 & 0.00  & 4.53 & 4.60 & 4.46 & 100.00 & 1.00 & 0.00 & 4.53 & 4.60 & 4.46 \\
    
  \end{tabular}
  }
  \caption{Quantitative comparisons of our model with state-of-the-art. We compare in two settings, a very low data setting (100 frames) and a somewhat low data setting (1000 frames). Our method is compared to person-specific models TalkLip \cite{wang2023seeing}, Wav2LipHQ \cite{Wav2LipHQ} and StyleSync \cite{guan2023stylesync}, and person-generic models including a baseline version of our model trained from scratch as well as GeneFace \cite{Ye2023GeneFaceGA} and RAD-NeRF \cite{tang2022radnerf}. We compare using quantitative (\textit{italics}) and user ratings (\textbf{bold}).}
  \label{tab:main_results}
\end{table*}

\subsection{Post Processing}
\label{sec:post}

While our method produces high-quality results in the facial interior, it occasionally suffers from artefacts around the border between the face and the background. Due to the strong bias introduced by the neural texture, pixels beyond the face region appear "stuck" to the face and follow its motion. This is best seen in video format, see the supplementary video for an example. In order to reduce the effect of this artefact, we apply post-processing. We first apply a semantic segmentation network \cite{yu2021bisenet} to each generated and real frame. This separates the face and neck from the background. We can then replace the generated pixels with the ground truth where both the generated and real frame agree the pixel is background.

\section{Results}

\textbf{Data:} Unless otherwise specified, we use the HDTF \cite{zhang2021flow} dataset. HDTF consists of around 400 videos, each in high-definition and several minutes long. We select a random subset of 20 of these videos to use for finetuning and testing, and the rest is used for the generic pertaining stage. We manually inspect the dataset to ensure that the subjects in the training set are not also accidentally included in the test set. We resample each video to 25fps and use 1500 frames (1 minute). We use the last 10 seconds of this as testing data and subsets of the remaining 50s for fine-tuning.

\noindent\textbf{Metrics:} We look to evaluate our method on three criteria, visual quality, lip sync and idiosyncracies. Visual quality is measured using FID. During any re-enactment experiments, as ground truth is available we also use SSIM and PSNR. While useful as proxies, these metrics are less important than how the method is perceived by users. For this reason, we also ask users to rate the three qualities: visual quality (QUAL), idiosyncracies (ID) and lip-sync (LIP) out of 5. A total of 30 users provided ratings. Further details of this user study are provided in the supplementary material.  

\subsection{Comparisons to State-of-the-Art}

\begin{figure}
    \centering
    \includegraphics[width=\columnwidth]{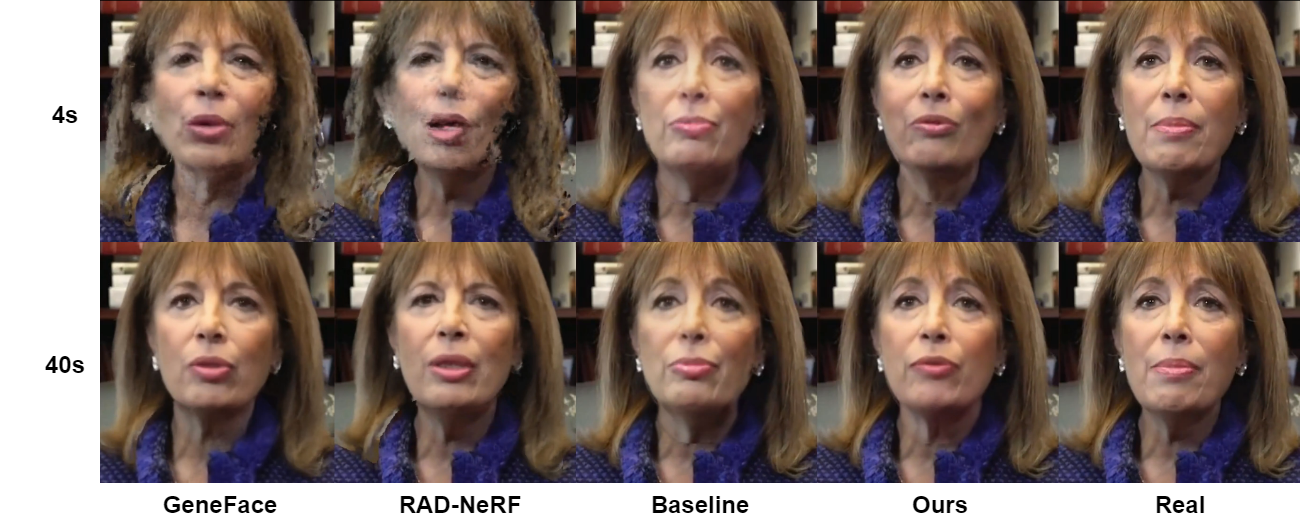}
    \caption{Qualitative comparisons to state-of-the-art person-specific models. We compare to GeneFace \cite{Ye2023GeneFaceGA}, RAD-NeRF \cite{tang2022radnerf} and the baseline model trained from scratch.}
    \label{fig:specific}
\end{figure}

We first compare our model to the existing state-of-the-art. We separate these models into person-specific and person-generic models. To demonstrate the ability of our model to work for both small and medium sized datasets, we consider two scenarios: one with very limited data (100 frames) and one with significantly more data (1000 frames).

We compare our model to three very recent person-generic methods: Wav2LipHQ \cite{Wav2LipHQ} which uses a VQ-GAN to achieve ultra-high resolution outputs, TalkLip \cite{wang2023seeing} which uses a lip reading network for better lip-sync, and the StyleGAN2 \cite{Karras2019stylegan2} based StyleSync \cite{guan2023stylesync}. For StyleSync, we use the generic version of the model due to the computation expense of training their specific model for every actor. We note that these papers have all compared to and significantly outperformed many older models, so we consider additional comparisons redundant. For person-specific models, we compare to the NERF-based RAD-NeRF \cite{tang2022radnerf} and GeneFace \cite{Ye2023GeneFaceGA}. We also compare to a baseline model. That is, we train our model from scratch on each subject. We consider this a close re-implementation of similar pipelines \cite{thies2020nvp, thies2019deferred}, with the relevant components upgraded. Therefore, we do not also compare our work to these models.

\begin{figure*}
    \centering
    \includegraphics[width=\textwidth]{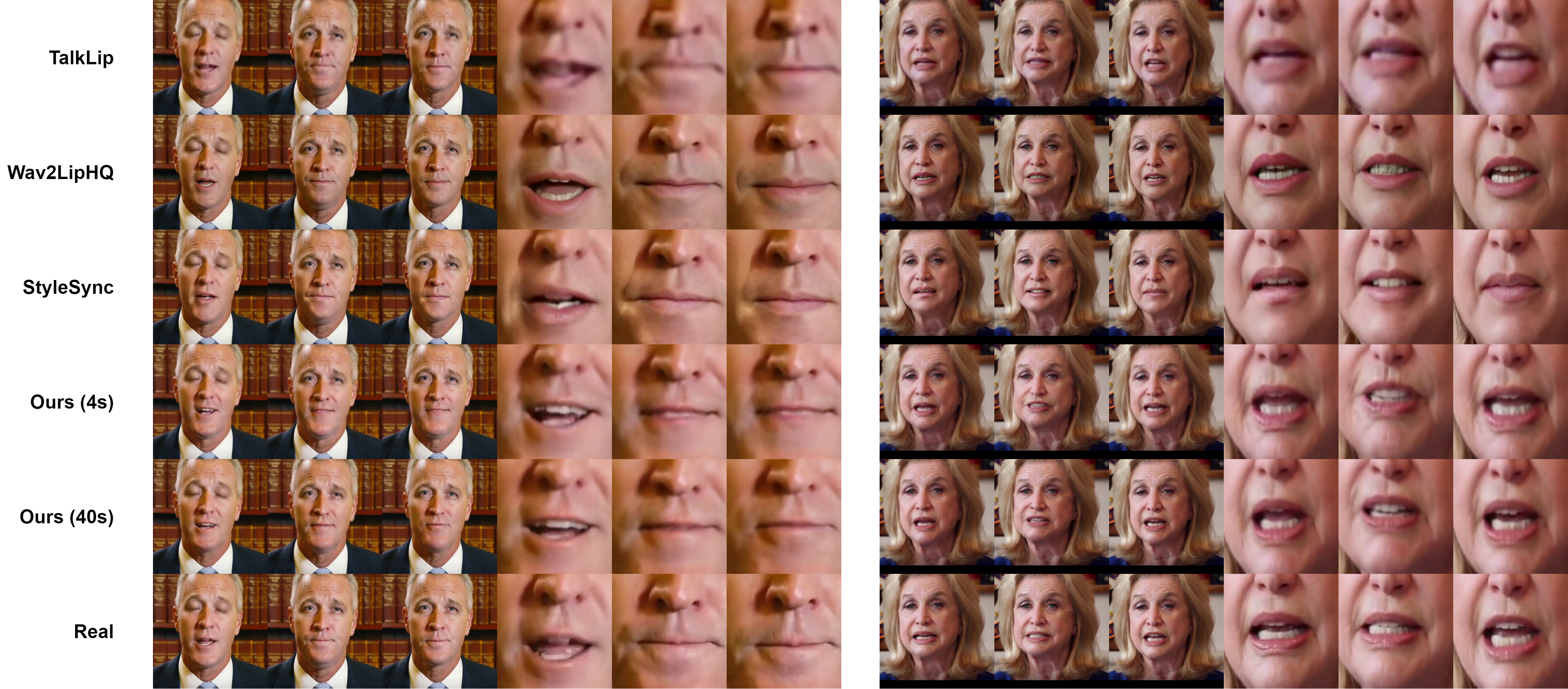}
    \caption{Qualitative comparisons to the state-of-the-art person generic models. We compare TalkLip \cite{wang2023seeing}, Wav2LipHQ \cite{Wav2LipHQ} and StyleSync \cite{guan2023stylesync} to our work. We show two versions of our model, one fine-tuned on 4 seconds of data and one with 40 seconds. We also show the ground truth frames for comparison. Closeups of the mouth region are included to better illustrate the results.}
    \label{fig:generic}
\end{figure*}

The results are shown in \Cref{tab:main_results}. \textbf{Our model outperforms the person-generic models in terms of visual quality as measured by FID}. This effect is more noticeable with additional data. User ratings for quality are slightly lower for our model. However, \textbf{our model is able to capture person-specific details that the generic models are not. This is evidenced in \Cref{fig:generic} and by the user ratings for idiosyncracies (ID)}. This may suggest that the generic models are producing very visually appealing but generic-looking lips. \textbf{When considering user ratings for lip-sync (LIP) our model performs slightly better.} Note that we do not use LSE metrics \cite{Wav2Lip} for lip sync as some previous works do. We find that the person-generic models optimise this directly, to the extent that they outperform the ground truth (9.34 for StyleSync vs 7.28 for real video) making this an unrelaible metric. Compared to person-specific models, our method outperforms them across all metrics. \textbf{The improvement is most prominent when trained on 4 seconds of data, suggesting our model is using the available data effectively}. We further investigate this effect in \Cref{sec:dataset_size}. The NeRF-based models in particular fail with unseen poses when trained on just four seconds of data. This can be seen easily in \Cref{fig:specific}. We also consider the version of our model in which we finetune only the texture and not the image-to-image network. Finetuning the full model increases the quality compared with tuning the textures only. This leaves a tradeoff between storage requirements and video quality that may be application-specific. For example, when dubbing a movie one might opt for high-quality tuning, whereas for translating millions of short-form videos, the texture only may be preferable.

\subsection{User Study}

\begin{table}[t]
    \centering
    \resizebox{\columnwidth}{!}{%
    \begin{tabular}{c|cccc}
        Row $>$ Col (\%) & Audio only & StyleSync & Baseline & Ours \\
        \hline
        Audio Only & - & 24 & 23 & 9 \\
        StyleSync & 76 & - & 68 & 38 \\
        Baseline & 77 & 32 & - & 9 \\
        Ours & 91 & 62 & 91 & -   
    \end{tabular}
    }
    \caption{User study performed on a translated section of an in-the-wild video clip. We show the percentage of 35 users who preferred the row to the column.}
    \label{tab:user_study}
\end{table}

\begin{table}[t]
    \centering
    \resizebox{\columnwidth}{!}{%
    \begin{tabular}{c|cccccc}
         & \multicolumn{6}{c}{Train Iterations to Reach PSNR} \\
        Method & 25 & 26 & 27 & 28 & 29 & 30 \\
        \hline
        Baseline &  2200 & 3000 & 6200 & 9800 & 18000 & 40000 \\
        Ours &  200 & 300 & 300 & 400 & 700 &  1600 \\
        \hline
        Speedup Factor & 11 & 10 & 20.7 & 24.5 & 25.7 & 25 \\
    \end{tabular}
    }
    \caption{Number of iterations (to the nearest 100) taken to reach a given PSNR value for our model and the baseline. Average of three runs with different identities.}
    \label{tab:speed}
\end{table}

\begin{table*}
  \centering
  \begin{tabular}{c|cccccc}

      & \multicolumn{6}{c}{HDTF 100 Frames}\\
     Method & PSNR $\uparrow$ & SSIM $\uparrow$ & FID $\downarrow$ & Qual $\uparrow$ & Lip $\uparrow$ & ID $\uparrow$   \\
    \midrule
    Ours w/o post-processing & 28.20 & 0.888 & \textbf{5.28} & 3.20 & 3.50 & 3.33 \\
    Ours w/o mixed data & 28.95 & 0.897 & 5.88 & 3.17 & 3.50 & 3.33 \\
    \hline
    \textbf{Ours full} & \textbf{29.10} & \textbf{0.899} & 5.76  & \textbf{3.57} & \textbf{3.77} & \textbf{3.83}\\
  \end{tabular}
  \caption{The results of the ablation study. We show that including our post-processing step (\Cref{sec:post}) and mixed training strategy (\Cref{sec:finetune}) both improve the results across many metrics.}
  \label{tab:ablation}
\end{table*}

To investigate our model in its intended context, altering the lip motion to match dubbed audio with limited data, we design a user study to replicate this. We take three videos of politicians speaking in their native language and use the automated (audio-only) dubbing provided by 11labs \cite{ElevenLabs}. These video clips are 15-20 seconds long, being much shorter than those used in previous works \cite{thies2020nvp, AudioDVP, Ye2023GeneFaceGA}. We compare our work to the highest-quality generic and specific models, measured by user ratings. We also consider audio-only dubbing (not altering the lips) which remains the industry standard. We perform a forced choice experiment. Users are given the same video dubbed using two random methods from our selection. The users are simply asked which they prefer. The results are in \Cref{tab:user_study} and show that users prefer our method to all others. Further details of the user study are outlined in the supplementary material.

\subsection{Ablations}

In this section, we perform an ablation study of our model. In particular, we show that the post-processing step (\Cref{sec:post}) and the mixed training strategy for fine-tuning (\Cref{sec:finetune}) both improve the results of our model. We use the 100-frame setting for this experiment. The results can be seen in \cref{tab:ablation}. The mixed training strategy improves results across all metrics, showing that it does indeed help the model generalise. The post-processing does increase FID, suggesting worse visual quality, however, the user ratings show a very clear preference for post-processing. This may be because the post-processing noticeably removes the sharp boundary between the real and generated frame, at the cost of some pixel-level accuracy.

\subsection{Training Speed}
\label{sec:speed}

Our model trains faster than existing person-specific models. This is due to the deferred neural rendering prior network. As this network works across many identities it requires much less training to adapt to a new identity. To demonstrate this effect, we compare our model to the baseline, a model trained from scratch for each new identity, which does not make use of the prior network and actor-specific adaptation. We show that our model is faster to train by recording the number of training iterations required to first reach a given value of PSNR on a withheld validation set. To show that this effect is not just due to limited data we train on the 1000 frame setting. The results are shown in \Cref{tab:speed} and show clearly that \textbf{our model trains an order of magnitude faster than existing models}.

\subsection{Effect of Dataset Size}
\label{sec:dataset_size}

Our method is designed to work for dubbing actors with only a few seconds of data. To demonstrate this ability we compare our method of using a prior network to a baseline model which simply trains both the deferred neural rendering network and neural texture from scratch. We evaluate the model using 10-second clips of one of our target actors, having trained both models on a subset of the training data. We plot the size of this subset against FID. The results are shown in \Cref{fig:dataset_size}, \Cref{fig:dataset_image} and in the supplementary video. It can be seen that \textbf{our model produces much higher quality results than the baseline on the small datasets}, but this effect reduces with dataset size.


\begin{figure}
    \centering
    \includegraphics[width=0.8 \columnwidth]{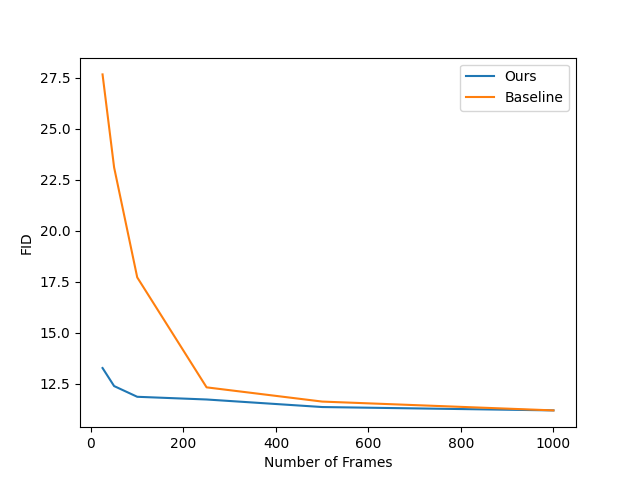}
    \caption{Our model is robust to small subsets of data. To show this we compare the amount of training data in frames to FID on a withheld validation set for a randomly selected actor.}
    \label{fig:dataset_size}
\end{figure}

\section{Limitations, Future Work and Ethics}

\begin{figure}
    \centering
    \includegraphics[width=\columnwidth]{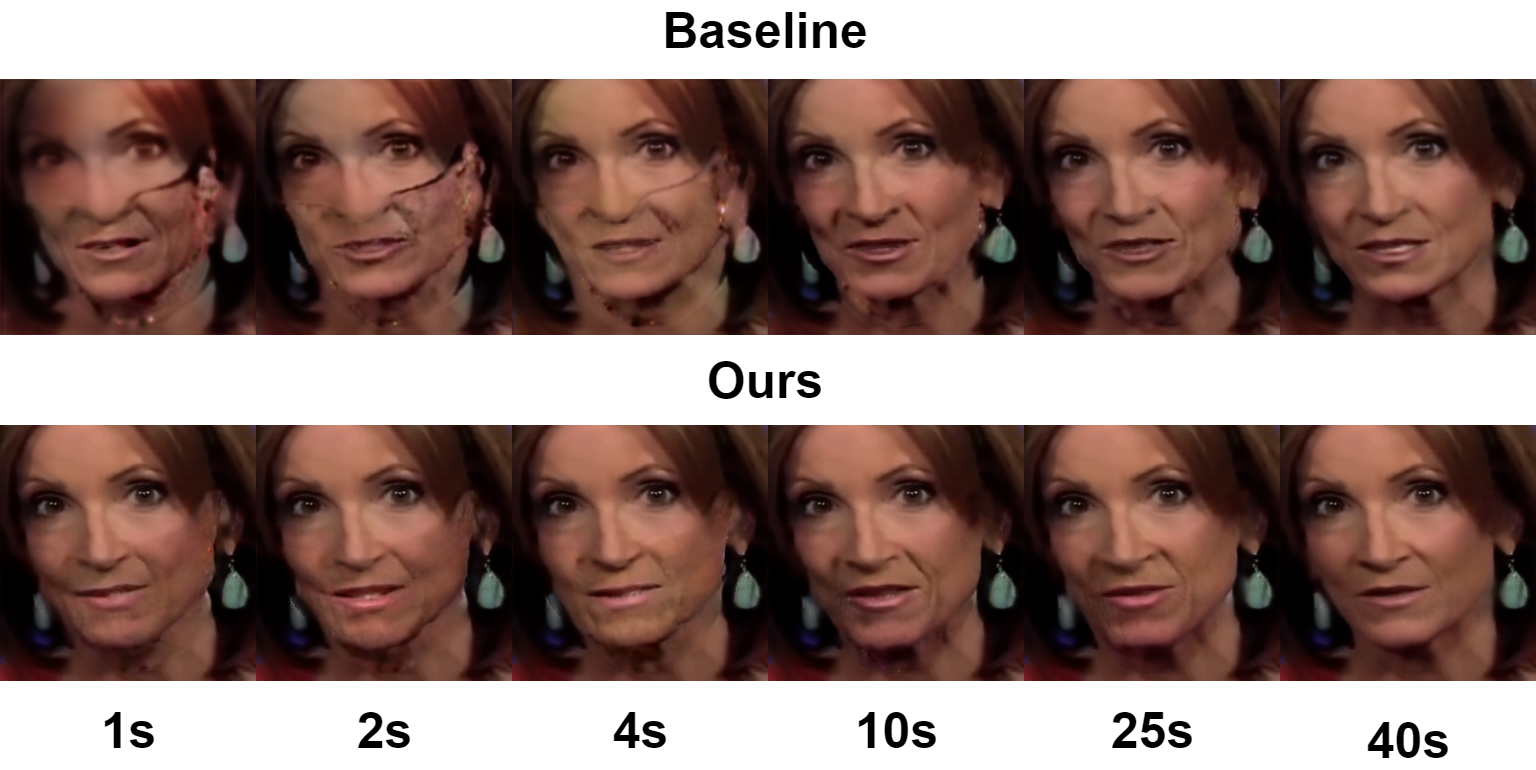}
    \caption{The effect of the size of the training dataset. The baseline model is trained from scratch and suffers greatly when using limited data. Ours, by comparison, is far more robust.}
    \label{fig:dataset_image}
\end{figure}

While our method achieves state-of-the-art, is robust to small datasets and trains fast, it is not without limitations. First, there are noticeable artefacts around the border between the face and the background. The post-processing we introduce does mitigate this, but not completely. We think this could be addressed by first segmenting the background from the person, training the model on the foreground only and composing the result. We will address this in future work. Another significant issue is that the monocular reconstruction stage is very slow as it relies on optimisation through a differentiable renderer. Recent work has shown \cite{DECA:Siggraph2021, EMOCA:CVPR:2021, filntisis2022visual} that regression-based reconstruction is possible, but it is still not temporally consistent enough for our purposes. We would like to investigate temporal regression models to this end, that could run in real-time.

\noindent \textbf{Ethical Considerations: } Our method has significant potential for misuse in creating misinformation and defamation. We believe strongly in managing the potential harms associated with this line of work and include a full ethics discussion in the supplementary.

\section{Conclusion}

We have presented Dubbing for Everyone. Our model is a hybrid person-generic, person-specific model which, using fine-tuning, is capable of high-quality visual dubbing using a few seconds of data for a given actor. Our experiments have shown that our model archives state-of-the-art across many metrics, including user ratings. We have also shown that our person-generic prior network training and adaptation strategy \textbf{trains faster, reaches higher quality and works on less data} than a similar model trained without priors.

{
    \small
    \bibliographystyle{ieeenat_fullname}
    \bibliography{main}
}

\clearpage
\setcounter{page}{1}
\maketitlesupplementary

\section{Further Methodological Details}

\subsection{Monocular Reconstruction}

Monocular reconstruction is the process of fitting a 3D mesh to the video. We follow a very similar pipeline to existing works \cite{MICA:ECCV2022, face2face, thies2020nvp, NHA}. That is, we use differentiable rendering using Pyorch3D to optimise a set of parameters to best fit the given video. We use the FLAME model \cite{FLAME:SiggraphAsia2017} to generate 5023 vertices from a set of parameters:

\begin{equation}
    V = F(\alpha, \beta, \theta)
\end{equation}

Where $\alpha \in \mathbb{R}^{300}$ are the parameters for shape, $\theta \in \mathbb{R}^{100}$ the expression parameters and $\phi$ the pose parameters for jaw, neck rotation and translation. We then model the image formation process using a perspective camera with intrinsics $K$ and extrinsics $R$, the PCA-based FLAME texture model with parameters $\beta$, and lighting (assumed distant and diffuse) using 9-band spherical harmonics with parameters $\gamma$. We optimise subsets of these parameters based on the following loss:

\begin{equation}
\label{recon_loss}
    \mathcal{L} = \lambda_{\text{photo}}\mathcal{L}_{\text{photo}} + \lambda_{\text{land}}\mathcal{L}_{\text{land}} + \lambda_{\text{reg}}\mathcal{L}_{\text{reg}}
\end{equation}

Where $\mathcal{L}_{\text{photo}}$ is the pixel-space $\ell_1$ loss between the rasterised image and the real frame. $\mathcal{L}_{\text{photo}}$ is the  $\ell_2$ distance between the projection of manually labelled landmarks on the FLAME mesh and corresponding landmarks detected with Mediapipe \cite{MediaPipe}. $\mathcal{L}_{\text{reg}}$ is a regularisation loss based on the $\ell_2$-norm of the shape, expression and pose parameters.

In stage 1 we predict $\alpha$ using the pre-trained MICA \cite{MICA:ECCV2022} model and do not change these after. In stage 2 we optimise \Cref{recon_loss} with respect to all parameters except $\alpha$ across several randomly selected frames. We then leave $\alpha, \beta, K, R$ all fixed and optimise \Cref{recon_loss} with respect to $\theta$ and $\phi$ for each frame sequentially starting at frame $0$. $\theta$ and $\phi$ are initialised at frame $t$ using their values at frame $t-1$. All optimisation is done with the Adam optimiser.

\subsection{Preprocessing}
In order to get a more stable bounding box for our preprocessing, we make use of the results of the monocular reconstruction. As the reconstruction is temporal and uses photometric losses, it is much more stable than bounding box estimation. To get a bounding box from the reconstructed mesh naively is simple. We simply produce the posed 3D mesh from the set of parameters using the FLAME model. We can then project the vertices of this mesh onto the image plane using the parameterised camera and take a bounding box that contains all the vertices plus some margin. 

Unfortunately, both this method and existing bounding box detectors have an issue related to the position of the jaw. When the jaw is open, the bounding box is longer than when it is closed. This may be desirable for some applications, but it is detrimental for our purposes. The reason is that the bounding box can bias the generation process. If the generator sees a longer box, it will infer that the jaw is open, so when trying to convert a frame with an open jaw to a closed one it will perform less well. We overcome this by making our bounding box jaw-independent. We set the jaw parameter of the FLAME model to fully closed, keeping everything else the same, and, project the vertices and then set it to fully open and do the same. We can then take the union of these points and get the bounding box from these. 

\section{Further Results}
\label{sec:user_studies}

\subsection{Two-Alternative Forced Choice User Study}

The Two-Alternative Forced Choice User Study (2AFC) is the choice of experiment in \Cref{tab:user_study}. For this experiment, we show the effectiveness of our method in the target scenario. We take a short video clip of approximately 15 seconds and use an online service to translate the audio with voice cloning. We do this for three videos taken from speeches, in various languages, from the European Central Bank YouTube page \cite{EUBank}.

We compare four methods. For our method, we finetune our pretrained model on just the 15 seconds of data. For the baseline model, we train the same model but from scratch, with all weights reinitialised. As StyleSync \cite{guan2023stylesync} scores best among the quantitative metrics besides our method, we use it for comparison. Finally, to show that altering the lips is actually necessary, we also consider the audio-only approach, where we keep the original video but change the audio stream.

To find out user preferences, we first show the user the real video and ask them to pay attention to the speaking style and appearance of the speaker. We then show two side-by-side, dubbed videos of the same person. These are selected randomly from all four methods and the order (left or right) is also random. We use Amazon's Mechanical Turk to collect these responses.

In total 35 users completed our user study. A total of 202 responses were recorded, meaning each user responded to an average of 5.8 comparisons. The vast majority (91\%) of users preferred our method to audio-only dubbing, suggesting that our method outperforms the industry standard. A similar percentage (91\%) preferred our method to the baseline. This is likely due to the relatively small size of the training dataset. Compared with StyleSync \cite{guan2023stylesync} a smaller, but still significant majority (61\%) preferred our work, suggesting that our model is indeed state-of-the-art for visual dubbing.

\subsection{Ratings User Study}

In order to compare our method to the existing state-of-the-art we also conduct a ratings user study. There are three qualities we wish to measure. These are: visual quality, which we denote QUAL in the results tables; idiosyncracies which we denote ID; and lip-sync which we denote LIP. We ask users to rate these three qualities out of 5, where 1 is very poor and 5 is very good, for each video and method. For visual quality, we ask users to consider artefacts and blur. For lip-sync we ask users to consider how well the lip movements match the audio. Finally, with idiosyncracies, we ask users to consider how much the video looks like the reference video, that is the real video that is being reconstructed, in terms of facial, lip, mouth and teeth appearance, as well as motion. The subject video and the reference are played separately to prevent poor lip-sync affecting the perceived idiosyncratic quality. 

Results are collected again using Mechanical Turk. In this case, each user is asked to rate all of the available methods in the same session, using the same subject. This allows the user to calibrate their ratings relative to all available methods. To prevent bias, the selected subject and the order the methods are shown to the user are randomised for every user. In total, 30 users completed the study, with each rating all methods exactly once. We report the mean opinion score in \Cref{tab:main_results} and  \Cref{tab:ablation}

\section{Further Ethical Discussion}

It is of paramount importance that the benefits and harms of the field of audio-driven visual dubbing are correctly weighted. In this section, we discuss these and detail our attempts to mitigate any harm.

\noindent \textbf{Bias: } The dataset we have used is the HDTF dataset. This dataset is collected from YouTube videos of politicians. It is worth noting that this dataset is biased in several ways. The authors do not provide demographic information, but it is clear that while the dataset does contain a range of ages, genders and ethnicities, it is biased towards white Americans. Importantly, the only language used in this dataset is English. We note, however, that this system is a prototype and we intend to further develop it using a much more diverse dataset before it may be used in any real-world settings.

\noindent\textbf{Privacy: } The HDTF dataset is realised under the Creative Commons Attribution 4.0 International License. We therefore have permission from the authors to use this dataset. Nonetheless, the original dataset collection methodology is unclear. To help protect the privacy of the individuals in this dataset and to comply with GDPR, we will provide a contact form allowing any of the individuals to remove themselves from the dataset and model. As our model is two-part, consisting of neural textures and a generic rendering network, the model is unable to reconstruct an individual without their neural texture. This means that simply deleting the texture will ensure that the individual is no longer represented in the model.

\noindent\textbf{Associated Harms: } The potential for our technology to be misused is significant. `Deepfakes" refers broadly to the class of artificially generated videos of which our work may be considered. These models are known to cause harm through misinformation, defamation and non-consensual explicit material, although our work does not enable the latter. To help mitigate these harms, we will only provide access to the model and its outputs to researchers at an accredited institution. We are actively investigating invisible watermarking \cite{watermarking} and deepfake detection methods \cite{roessler2019faceforensicspp, mirsky2021creation}.

\noindent\textbf{Associated Benefits: } Our model enables media to cross language barriers. This helps promote diverse societies and allows for the spread of various cultures. In addition to this, the method has significant potential economic value. In these ways, the development of such models can benefit societies. Furthermore, by providing access to our model to researchers working on deepfake detection and video verification, we help support these goals that may otherwise fall short of bad actors developing similar models to ours. 

The exact weighting of the good and harms of developing ``deepfake" models remains an open question, but we believe that visual dubbing, with its potential for the spread of culture and the economic benefits, outweigh the potential risks when considering the mitigation we have put in place.

\end{document}